\documentclass{article}

\usepackage{arxiv}

\usepackage[utf8]{inputenc}
\usepackage[T1]{fontenc}
\usepackage{hyperref}
\usepackage{url}
\usepackage{booktabs}
\usepackage{amsfonts}
\usepackage{nicefrac}
\usepackage{microtype}
\usepackage{xcolor}
\usepackage[numbers, compress]{natbib}

\usepackage{graphicx}
\usepackage{wrapfig}
\usepackage{amsmath} 
\usepackage{longtable}
\usepackage{multirow}

\title{Towards Long-Context Time Series Foundation Models}

\author{
  Nina Żukowska \quad Mononito Goswami \quad Michał Wiliński \quad Willa Potosnak \quad Artur Dubrawski \\
  Auton Lab, School of Computer Science, Carnegie Mellon University\\
  Pittsburgh, PA 15213\\
  \texttt{\{nzukowsk,mgoswami\}@andrew.cmu.edu} \\
}

\begin{document}
\maketitle

\begin{abstract}
Time series foundation models have shown impressive performance on a variety of tasks, across a wide range of domains, even in zero-shot settings. However, most of these models are designed to handle short univariate time series as an input. This limits their practical use, especially in domains such as healthcare with copious amounts of long and multivariate data with strong temporal and intra-variate dependencies. Our study bridges this gap by cataloging and systematically comparing various context expansion techniques from both language and time series domains, and introducing a novel compressive memory mechanism to allow encoder-only TSFMs to effectively model intra-variate dependencies. We demonstrate the benefits of our approach by imbuing \texttt{MOMENT}, a recent family of multi-task time series foundation models, with the multivariate context.
\end{abstract}

\section{Introduction}

Large Language and Vision Models (LLMs and LVMs) have revolutionized text and image modeling, enabling a wide range of applications with both limited data and expert supervision. Time series foundation models (TSFMs)~\cite{goswami2024moment, garza2023timegpt1, ansari2024chronos, rasul2023lag, woo2024unified, ekambaram2024ttms, das2023decoder, liutimer} promise to bring similar transformative advancements to modeling time series. However, most of these models, barring \texttt{MOIRAI}~\cite{woo2024unified} and \texttt{TTMs}~\cite{ekambaram2024ttms}, can only model short univariate time series, limiting their widespread use in applications such as healthcare where long and multivariate time series are common. Most of these approaches downsample long time series to handle extended context lengths and model different channels independently to manage multivariate inputs, which limits their ability to capture potentially informative high-frequency and intra-variate dependencies.

A straightforward solution to modeling both long and multivariate inputs is to re-design and pre-train TSFMs with longer context lengths and to concatenate multiple channels sequentially~\cite{woo2024unified}. However, this naive solution drastically increases computational complexity as Transformer-based foundation models~\cite{goswami2024moment, garza2023timegpt1, ansari2024chronos, rasul2023lag, woo2024unified, das2023decoder, liutimer} are constrained by context-dependent memory, due to their quadratic complexity in the length of the input. Recent studies have explored the use of compressive memory to enable Transformer-based LLMs to process very long sequences with bounded memory and computation~\cite{munkhdalai2024leave}. We adapt these techniques to design a novel compressive memory mechanism which we call \textbf{Infini-Channel Mixer (ICM)}, that can allow encoder-only Transformers~\cite{goswami2024moment, woo2024unified, nie2023patchtst} to efficiently model intra-variate dependencies. We use Infini-Channel Mixer to imbue \texttt{MOMENT}, a recent family of multi-task TSFMs, with multivariate context. 

Our contributions include: \textbf{(1)} We outline the \textit{design space} of context expansion techniques from both language and time series modeling; \textbf{(2)} We propose Infini-Channel Mixer, a \textit{novel compressive memory mechanism} to enable encoder-only Transformers to model multivariate time series; and \textbf{(3)} We systematically compare various context expansion techniques with increasing levels of complexity on multivariate long-horizon forecasting to demonstrate that our proposed approach can improve forecasting performance and efficacy for the  \texttt{MOMENT}~\cite{goswami2024moment} TSFM family.  

\section{Related Work}
\subsection{Foundation Models} Foundation models have significantly influenced several domains, particularly natural language processing and computer vision. These models typically rely on large-scale pre-training on vast, unlabeled datasets. For example, \texttt{GPT} models \citep{radford2019language, brown2020gpt3} introduced autoregressive language modeling, while \texttt{BERT} \citep{devlin2018bert} and \texttt{RoBERTa} \citep{liu2019roberta} popularized masked language modeling with Transformer-based architectures. Most of these models are Transformer-based and designed for long-context tasks, with typical context lengths between 512 and 5000 tokens. The encoder-only architectures like \texttt{BERT} excel at representation learning, whereas autoregressive models such as \texttt{GPT} perform well in generative tasks.

\subsection{Time Series Foundation Models} The domain of time series analysis has also seen the rise of foundation models, which share many similarities with NLP models, particularly in their use of Transformer architecture. \texttt{TimeGPT} \citep{garza2023timegpt1} was the first time series foundation model, setting a foundation for using pre-training in time series forecasting. \texttt{Time-LLM} \citep{jin2023time} demonstrated the potential for adapting LLMs to time series tasks. \texttt{PatchTST} \citep{nie2023patchtst} introduced the concept of patching time series data, enabling the models to handle longer sequences efficiently. \texttt{MOMENT} \citet{goswami2024moment}, stands out for its ability to solve multiple time series tasks, within a unified framework. \texttt{MOMENT} is open-source, with access to training data and code, making it accessible for a wide range of applications. 

 Our work builds on the advances of these foundation models, particularly leveraging the flexible, multitask capabilities of \texttt{MOMENT}. Most existing time series foundation models are Transformer-based, with either encoder-only or autoregressive architectures. Our proposed Infini-Channel Mixer extends the encoder-only architecture to better handle multichannel data, improving performance on time series forecasting tasks while maintaining model simplicity.
\section{Design Space of Multivariate Time Series Models}
\label{sec:design_space}
\begin{wrapfigure}[17]{r}{0.6\textwidth}
    \includegraphics[width=\linewidth]{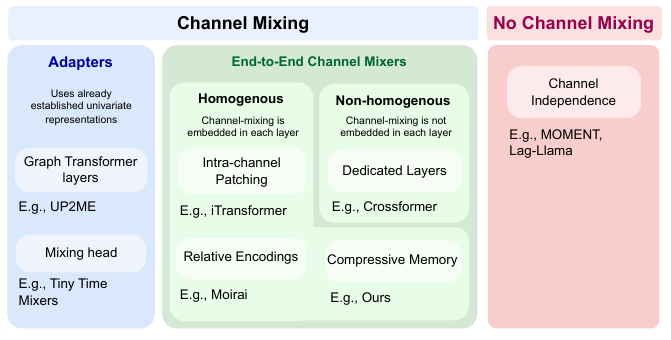} 
    \caption{A design space of multivariate time series models. The proposed Infini-Channel Mixer is a homogeneous end-to-end channel mixing method.}
    \label{fig:taxonomy}
\end{wrapfigure}

\textbf{Channel Independence.} Many multivariate time series models \cite{nie2023patchtst}, including foundation models~\cite{goswami2024moment, rasul2023lag, das2023decoder, ansari2024chronos}, treat different channels independently. These approaches are simple, scalable, and typically perform well on real-world data and academic benchmarks~\cite{zhou2021informer} without substantial intra-channel dependencies.

To capture informative intra-channel dependencies in real-world time series, various \textit{channel-mixing} approaches have been proposed in the literature:

\textbf{Adapters.} Approaches in this family usually first learn representations for each channel using a univariate backbone and then combine them using a multi-channel adapter. Examples include multivariate decoders~\cite {ekambaram2024ttms} and Graph Transformer layers~\cite{zhangup2me}.

\textbf{End-to-End Channel Mixers.} These approaches tightly integrate channel mixing throughout the architecture. For instance, \texttt{MOIRAI}~\citep{woo2024unified} flattens channels and uses relative channel encodings to distinguish information from different channels. \texttt{iTransformer}~\citep{liu2024itransformer} attends to the inverted ``variate`` tokens which capture multi-variate correlations. In each of these approaches, every layer of the model is \textit{homogeneous} and performs the same computation. Other methods such as \texttt{Crossformer}~\citep{zhang2023crossformer} are \textit{non-homogeneous} as they use different attention matrices to model inter-sequence and intra-channel information. Our proposed Infini-Channel Mixer method attaches compressive memory to each Transformer layer and is therefore a homogeneous end-to-end channel mixer by design.
\newpage
\section{Infini-Channel Mixer: Unlocking Multivariate Context using Compressive Memory}
\label{headings}
\begin{wrapfigure}[22]{l}{0.5\textwidth}
    \includegraphics[width=\linewidth]{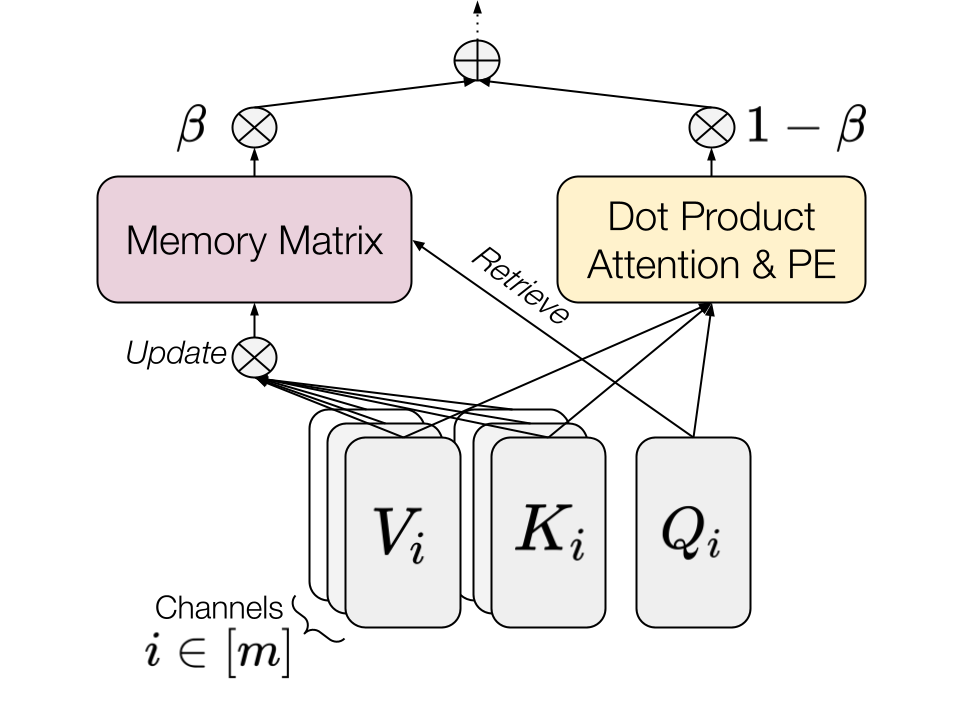}
    \caption{Infini-Channel Mixer (ICM) uses a learned scalar $\beta$, to balance local information from dot product attention with global information from the compressive memory matrix which aggregates cross-channel information.}
    \label{fig:infini_channel_mixer}
\end{wrapfigure}

\textbf{Requirements.} Our goal is to modify the standard transformer architecture minimally to accommodate multivariate time series. To simplify development and keep multivariate context expansion our primary focus, we fix the maximum length of time series that can be processed by our model. Given the prevalence of encoders in time series foundation models~\cite{goswami2024moment, woo2024unified, ansari2024chronos}, the proposed architecture should be amenable to bidirectional self-attention. 

\textbf{Infini-Channel Mixer (ICM).} To aggregate information from an arbitrary number of channels, we propose incorporating compressive memory into Transformers. Compressive memory systems use a fixed number of parameters to store and retrieve information efficiently. Our approach is inspired by Infini-attention \citep{munkhdalai2024leave}, a recent compressive memory-based attention mechanism that has enabled \textit{decoder-only} LLMs to, in theory, scale to infinitely long input sequences. Unlike Infini-attention, our approach uses memory to process multiple channels in transformers with encoders.

Infini-Channel Mixer extends a standard self-attention layer with a \textit{compressive memory matrix} allowing it to attend to both local (intra-channel) and global (inter-channel) information. The memory matrix is computed by reusing the key (\( \mathbf{K} \)), query (\( \mathbf{Q} \)), and value (\( \mathbf{V} \)) matrices from the dot product attention as shown in Fig.~\ref{fig:infini_channel_mixer}, which accelerates training and inference and acts as a regularizer. A learned gating scalar $\beta$ regulates the trade-off between global information queried from the memory matrix and local information from the dot-product attention of a specific channel. This memory mechanism is replicated for each attention head and layer.

\textbf{Step 1: Aggregate Cross-channel Information into Compressive Memory.} The compressive memory matrix, \( \mathbf{M} \in \mathbb{R}^{h \times d_k \times d_k} \), and the normalization term, \( \mathbf{z} \in \mathbb{R}^{h \times d_k \times 1} \), are initially set to zero. Here, \( h \) is the number of attention heads, and \( d_k \) is the key dimension. We store information from each channel, $i \in [m]$, by reusing its local \( \mathbf{KV} \) entries. To ensure training stability, we follow prior work, and set the normalization term to the sum of key entries from all channels \cite{munkhdalai2024leave, pmlr-v119-katharopoulos20a}, and use element-wise (Exponential Linear Unit) ELU $+\ 1$ as a non-linear projection function \(\sigma\). We define \textit{n} as sequence length and $\mathbf{K}^{i}_{j}$ denotes the key entries of the $i^{\text{th}}$ channel and the $j^{\text{th}}$ input token.
\begin{equation} \label{eq:M_z_update}
\begin{aligned}
\mathbf{M} &\leftarrow \mathbf{M} + \sigma(\mathbf{K}^i)^{\top} \mathbf{V}^i \hspace{20pt}
\mathbf{z} & \leftarrow \mathbf{z} + \sum_j^{n} \mathbf\sigma(\mathbf{K}^{i}_{j})
\end{aligned}
\end{equation}

\textbf{Step 2: Conditioning on Inter-channel and Intra-Channel Information.} We use the query matrix to retrieve the cross-channel information \(\mathbf{A}_{\text{mem}^i}\) from the memory. This retrieved information is then combined with the local attention state \(\mathbf{A}_{\text{dot}^i}\) using the learned gating scalar $\beta$. This design adds only a single trainable parameter per head to regulate the trade-off between local and global information.
\begin{equation} \label{eq:M_z_query}
\begin{aligned}
\mathbf{A}_{\text{mem}^i} = \frac{\sigma(\mathbf{Q}^i)\mathbf{M}}{\sigma(\mathbf{Q}^i)\mathbf{z}  + \epsilon}\hspace{20pt
}\mathbf{A}_{i} = sigmoid(\beta) \odot \mathbf{A}_{\text{mem}^i} + (1 - sigmoid(\beta)) \odot \mathbf{A}_{\text{dot}^i}
\end{aligned}
\end{equation}
\section{Experiments and Results}
The goals of our experiments are twofold: (1) to systematically compare ICM against alternative context expansion techniques, and (2) to assess its impact on pre-training. To evaluate context expansion techniques outlined in Section~\ref{sec:design_space}, we conduct several small-scale supervised experiments focusing on the long-horizon forecasting task. Then to study the impact of ICM on pre-training, we pre-train the same model both with and without ICM and then evaluate the resulting models on two downstream tasks: long-horizon forecasting and multivariate classification. 
\newpage

\begin{table}[h!]
\centering
\resizebox{\textwidth}{!}{
\begin{tabular}{r|cc|ccccccc}
\toprule
 \textbf{Model / Example} & \textbf{Class} & \textbf{Design} & \texttt{Exchange} & \texttt{ETTh1} & \texttt{ETTh2} & \texttt{ETTm1} & \texttt{ETTm2} & \texttt{Weather} \\ \midrule
N-BEATS \citep{oreshkin2019nbeats} & \multirow{2}{*}{\begin{tabular}[c]{@{}c@{}}No Channel\\Mixing\end{tabular}} & \multirow{2}{*}{\begin{tabular}[c]{@{}c@{}}Channel\\Independence\end{tabular}} & 0.524 & 0.461 & 0.410 & 0.346 &  0.278 & 0.211 \\
\texttt{MOMENT-Tiny} \cite{goswami2024moment} &  &  & 0.249 & \underline{0.418} & 0.359 & \underline{0.339} & \textbf{0.234} & 0.206 \\ \midrule
UP2ME \citep{zhangup2me} & Adapter & Graph Transformer & \underline{0.240} & 0.435 & 0.367 & 0.340 & \underline{0.237} & \textbf{0.204} \\ \midrule
Crossformer \citep{zhang2023crossformer} & \begin{tabular}[c]{@{}c@{}}Non-homogeneous\\End-to-End Mixer\end{tabular} & \begin{tabular}[c]{@{}c@{}}Dedicated Intra-\\Channel Attention\end{tabular}  & 0.559 & 0.571 & 0.654 & 0.390 & 0.515 & 0.227 \\ \midrule
iTransformer \cite{liu2024itransformer} & \multirow{4}{*}{\begin{tabular}[c]{@{}c@{}}Homogeneous\\End-to-End\\Channel Mixer\end{tabular}} & Multivariate Patching & 0.245 & 0.429 & 0.380 & 0.353 & 0.251 & 0.212 \\
MOIRAI \citep{woo2024unified} &  & \begin{tabular}[c]{@{}c@{}}Concatentation\\$+$ Relative Encoding\end{tabular} & 0.243 & 0.426 & \underline{0.357} & 0.340 & 0.249 & 0.216 \\ 
ICM (Ours) &  & Compressive Memory & \textbf{0.232} & \textbf{0.416} & \textbf{0.349} &  \textbf{0.333} & \textbf{0.234} & \underline{0.205} \\

\bottomrule
\end{tabular}
}
\caption{Forecasting MSE of all models averaged over 3 different horizons $[96, 192, 384]$. The best models are shown in \textbf{bold} and the second best ones are \underline{underlined}. In most cases, Infini-Channel Mixer (ICM) outperforms considered alternatives. Complete results in Table~\ref{tab:metrics_forecast_pretrained_full} (Appendix~\ref{appendix:results}).}
\label{tab:metrics_forecast_supervised_subset}
\end{table}

\vspace{-4mm}

\paragraph{Supervised Long-horizon Forecasting Experiments.} Due to variations in architectures and experimental setups, findings from prior work are inconclusive on how to best endow multivariate context to time series models. We address this by carefully implementing all context expansion techniques from Figure~\ref{fig:taxonomy} on the same model architecture and experimental settings of long-horizon forecasting~\cite{zhou2021informer}. We use \texttt{MOMENT}, a recent open-source multi-task foundation model as the base architecture for all our experiments. To accelerate the process, we introduce a \texttt{Tiny} variant of it based on \texttt{T5-Efficient-Tiny} architecture \cite{tay2022scaleefficientlyinsightspretraining}. All model variants take time series of length 256 as input (lookback window) and forecast the next 96, 192, and 384 time steps. As is common practice, models are trained and evaluated using Mean Squared Error (MSE). The exact details of model architecture, experimental settings, and hyper-parameters are outlined in Appendix~\ref{appendix:setup}.

Results summarized in Table~\ref{tab:metrics_forecast_supervised_subset} show that our proposed approach outperforms alternative context expansion techniques on average in most cases. 

\begin{table}[th!]
\centering
\resizebox{0.8\textwidth}{!}{
\begin{tabular}{rc|cccccc}
\toprule
\textbf{Model name}  & Fine-tune $\beta$ & \texttt{Exchange} & \texttt{ETTh1} & \texttt{ETTh2} & \texttt{ETTm1} & \texttt{ETTm2} & \texttt{Weather} \\ \midrule
\texttt{MOMENT-Tiny} & $-$ & 0.250 & \underline{0.437} & 0.343 & 0.333 & \underline{0.230} & 0.222 \\ \midrule
\multirow{2}{*}{\begin{tabular}[c]{@{}c@{}}+Infini-Channel\\Mixer\end{tabular}} & $\times$ &\underline{ 0.249} & 0.439 & \textbf{0.336} & \underline{0.332} & \underline{0.230} & \underline{0.219} \\
& $\checkmark$ & \textbf{0.247} & \textbf{0.436} & \underline{0.337} & \textbf{0.330} & \textbf{0.228} & \textbf{0.214} \\
\bottomrule
\end{tabular}
}
\caption{Forecasting MSE averaged over 3 horizons $[96, 192, 384]$. \texttt{MOMENT-Tiny} with Infini-Channel Mixer outperforms its vanilla variant on all datasets. Additionally, fine-tuning the $\beta$ parameters along with the forecasting head most often results in improved performance.}
\label{tab:impact_of_beta}
\end{table}

\vspace{-4mm}
\begin{wraptable}[12]{r}{0.5\textwidth}
\centering
\small
\centering
\begin{tabular}{r|cc}
\toprule
{} & \texttt{MOMENT-Tiny} & \texttt{+ ICM}\\
\midrule
Mean  & 0.625 & 0.632 \\
Std.   & 0.254 & 0.268 \\
Median  & 0.616 & 0.704 \\
Wins/Ties/Losses & 12/2/12 & 12/2/12 \\
\bottomrule
\end{tabular}
\caption{Accuracy of \texttt{MOMENT-Tiny} with and without ICM across 26 UEA multivariate classification datasets. Complete results in Table~\ref{table:multivariate_classification}. The addition of compressive memory does not substantially change multivariate classification accuracy, however it improves results on smaller datasets.}
\label{table:summary_stats}
\end{wraptable}

\paragraph{Pre-training Experiments.} We pre-train 2 variants on \texttt{MOMENT-Tiny}: one with Infini-Channel Mixer and one without it. Pre-training multivariate foundation models is challenging due to the scarcity of public benchmark multivariate datasets and the complexity of managing time series with varying numbers of channels. To address these issues, we pre-train our models for one epoch on the Time Series Pile~\cite{goswami2024moment}, using a mix of univariate and multivariate datasets. To manage memory consumption during training, we fix the maximum number of channels ($= 8$) per batch and sub-sample time series with more channels. Both models are trained with a maximum sequence length of 256 time steps. 

We evaluate the pre-trained models on two downstream tasks: long-horizon forecasting and unsupervised representation learning for classification. For the forecasting task, we fine-tune a simple linear forecasting head on the representations learned by our models, while keeping all other model parameters frozen, except for those in the forecasting head and the $\beta$ coefficients. We use the same 6 datasets as in the previous experiment. For classification, following standard practice \cite{goswami2024moment}, we obtain representations of time series in a zero-shot setting (without access to labels) and use these representations to train a Support Vector Machine. For these experiments, we use the \texttt{UCR/UEA} classification repository \cite{UCRArchive} which comprises both univariate and multivariate datasets, frequently used to benchmark classification algorithms. 

Tables~\ref{tab:impact_of_beta} and~\ref{table:summary_stats} summarize forecasting and classification results, details in Appendix~\ref{appendix:results}.

\paragraph{Discussion.} Our experiments demonstrate the potential of the Infini-Channel Mixer as an effective context expansion mechanism. Despite adding only 16 new parameters (one for each of the 16 attention heads in \texttt{MOMENT-Tiny}), our approach consistently outperformed alternative methods on datasets with cross-channel dependencies. Interestingly, on some datasets like \texttt{ETTm2} and \texttt{Weather}, treating channels independently yielded better performance than many multivariate methods. This finding suggests that current academic benchmarks might not fully capture the benefits of multivariate modeling. The promising performance of the \texttt{UP2ME} \cite{zhangup2me} approach, which models a subset of channels with substantial cross-correlation, supports this observation. Additionally, the substantial performance gains achieved by fine-tuning the $\beta$ coefficients further indicate that they function as a soft-gating mechanism, enabling the model to focus on the most informative cross-channel information.
\section{Conclusion and Future Work}
We introduced Infini-Channel Mixer (ICM) , a novel compressive memory mechanism that enables encoder-based TSFMs to effectively handle multivariate time series data. We systematically compared our proposed approach with multiple context expansion techniques proposed in the literature. We demonstrated that ICM, with only a few additional parameters, can improve the performance of time series models on two important tasks: long-horizon forecasting and classification. Future work should focus on rigorously evaluating our approach on larger models, across a wider range of datasets and tasks. Additionally, there is a pressing need for more multivariate datasets with strong cross-channel dependencies to better pre-train and evaluate large-scale models. It will also be important to explore whether Infini-Channel Mixer can be effectively adapted to handle extremely long time series.

\bibliographystyle{plainnat}
\bibliography{references}

\appendix
\section{Channel Mixing}
\subsection{Concatenation Approach} 
\label{appendix:naive-mixer}
In order to perform information exchange between the multiple variates, they have to be processed either at the same time, producing a higher memory cost, or sequentially, producing a higher computational cost. The default way of processing multiple variates is related to flattening the data from all channels, concatenating it, and producing an attention matrix with two dimensions equal to (\textit{channel number $\times$ sequence length}). \citet{woo2024unified} implements a similar approach with inter- and intra-channel bias scalars, akin to the concept of relative positional biases in \citet{raffel2020exploring}. The approach of concatenating \citep{woo2024unified} all of the variates is expressed in \ref{eq:baseline_approach}. To be able to ablate the approach, we do not apply the proposed rotary positional encoding, we stick to the sinusoidal position encoding.

\begin{equation} \label{eq:baseline_approach}
E_{ij,mn} = \left( \mathbf{x}_{i,m} \mathbf{W}^Q \right) \left( \mathbf{x}_{j,n} \mathbf{W}^K \right)^\top + \delta_{mn} \cdot u^{(1)} + (1 - \delta_{mn}) \cdot u^{(2)}
\end{equation}

\noindent
where:
\begin{itemize}
    \item \(\mathbf{x}_{i,m}\) and \(\mathbf{x}_{j,n}\) are the input vectors for tokens \(i\) and \(j\) of variates \(m\) and \(n\), respectively.
    \item \(\mathbf{W}^Q\) and \(\mathbf{W}^K\) are the weight matrices for queries and keys.
    \item \(u^{(1)}\) and \(u^{(2)}\) are scalar bias terms. \(u^{(1)}\) is added when \(m = n\), and \(u^{(2)}\) is added when \(m \ne n\).
    \item \(\delta_{mn}\) is the Kronecker delta function, which is 1 if \(m = n\) and 0 otherwise.
    \item \(E_{ij,mn}\) represents the attention score between tokens \(i\) and \(j\) of variates \(m\) and \(n\), respectively.
\end{itemize}

\subsection{Infini-Channel Mixer + Static}
\label{appendix:static_channel_embedding}
We would like to point out, that the novel concept of a memory matrix opens up possibilities to query a matrix differently, based on the inter-channel relations. A simple, but naive way of ensuring that is to add channel embeddings to each patch embedding in the embedding layer.

Let \( \mathbf{E}_{\text{channel}} \in \mathbb{R}^{\text{channel\_num} \times d_{\text{model}}} \) be the matrix of static channel embeddings, where each row represents a unique embedding for a channel.
The channel embeddings are broadcasted to match the dimensions of the input tensor and added to the patch embeddings. This operation can be mathematically represented as:

\begin{equation}
\mathbf{x}'_{b, c, t, d} = \mathbf{x}_{b, c, t, d} + \mathbf{E}_{\text{channel}, c, d}
\end{equation}

where:
\begin{itemize}
    \item \(\mathbf{x}_{b, c, t, d}\) is the patched and embedded input
    \item \(\mathbf{E}_{\text{channel}, c, d}\) is the learned static channel embedding for the \(c\)-th channel and \(d\)-th dimension,
    \item \(\mathbf{x}'_{b, c, t, d}\) is the updated embedding after adding the channel embedding.
\end{itemize}

This method is denoted as "Infini-Channel Mixer + Static" in the supervised setting in the table \ref{tab:metrics_forecast_supervised_full}

\section{Training Experimental Setup}
\label{appendix:setup}
\subsection{Supervised Setting}

In the supervised setting, we evaluate the following models using a consistent training setup:

\begin{itemize}
    \item \textbf{N-BEATS:} For training \texttt{N-BEATS}, we use the following configuration: Stack Types (Trend and Seasonality), Number of Blocks per Stack (3), Theta Dimensions (4 and 8), and Hidden Layer Units (256). The training is conducted for 10 epochs with a batch size of 64, and is directly copied from \citet{goswami2024moment}.
    
    \item \textbf{Concatenation and Infini-Channel Mixer:} For the Concatenation, Infini-Channel Mixer, and Infini-Channel Mixer + Static models, we employ the \texttt{T5-Efficient-TINY} backbone. This backbone comprises 4 encoder blocks, a model dimension of 256, 4 attention heads, and feed-forward dimensions of size 1024. Detailed specifications for the concatenation method are included in the appendix \ref{appendix:naive-mixer}.
    
    \item \textbf{MT + Graph Transformer Layer:} For the MT model with an added Graph Transformer layer, we use the \texttt{MOMENT-Tiny} backbone, incorporating a Graph Transformer layer as detailed in \citet{zhangup2me}.
    \item \textbf{Crossformer:} We use the original parameters of the \texttt{Crossformer} including patch size 12, 3 encoder block layers, the feed-forward dimension of 128, model dimension of 256, and 4 attention heads.
\end{itemize}

\section{Forecasting and classification task}
\label{appendix:results}

\paragraph{Pre-training} We use the pre-trained models: \texttt{MOMENT-Tiny}, \texttt{MOMENT-Tiny + Infini-Channel Mixer}. Both are pre-trained for one epoch and with the same amount of data. We want to ensure a constant number of channels in a single batch (either 1 or 8), thus for multivariate datasets, we divide them into subdatasets of 8 channels. In case we end up with less than 8 channels in a subdataset, we oversample the remaining channels. This yields two types of batches -  univariate and multivariate. We then use them in the pre-training.

\subsection{Forecasting}
Since  \texttt{MOMENT} can handle multiple tasks, we test our approach on forecasting (supervised \ref{tab:metrics_forecast_supervised_full} and pre-trained \ref{tab:metrics_forecast_pretrained_full}) and classification tasks (multivariate and univariate pre-trained models). After pre-training the \texttt{MOMENT-Tiny + Infini-Channel Mixer} models, we fine-tune the linear head, as well as the $\beta$ parameters.

\begin{table}[h!]
\centering
\resizebox{\textwidth}{!}{
\begin{tabular}{r|c|cc|cc|cc|cc|cc|cc}
\toprule
\textbf{Model name} & \textbf{Horizon} & \multicolumn{2}{c}{\texttt{Exchange}} & \multicolumn{2}{c}{\texttt{ETTh1}} & \multicolumn{2}{c}{\texttt{ETTh2}} & \multicolumn{2}{c}{\texttt{ETTm1}} & \multicolumn{2}{c}{\texttt{ETTm2}} & \multicolumn{2}{c}{\texttt{Weather}} \\
& & MAE & MSE & MAE & MSE & MAE & MSE & MAE & MSE & MAE & MSE & MAE & MSE \\
\midrule
Pre-trained \texttt{MT + Infini-Channel Mixer} (8 folds) & 96  &  \underline{0.229} &  \underline{0.104} &  \textbf{0.407} &  \underline{0.403} &  \textbf{0.344} &     \textbf{0.288} &  \textbf{0.347} &  \underline{0.293} &     \textbf{0.257} &  \textbf{0.171} &   \underline{0.217} &  \underline{0.166} \\
& 192 &     \textbf{0.326} &     \textbf{0.207} &  \textbf{0.427} &  \underline{0.438} &  \textbf{0.388} &     \textbf{0.351} &  \textbf{0.367} &     \textbf{0.327} &     \textbf{0.295} &  \textbf{0.224} &      \textbf{0.252} &  \underline{0.207} \\
& 384 &              0.482 &              0.430 &           0.447 &              0.467 &           0.413 &     \textbf{0.371} &  \textbf{0.391} &  \underline{0.369} &              0.340 &  \textbf{0.291} &                  \underline{0.299} &  \underline{0.271} \\
\midrule
Pre-trained \texttt{MT + Infini-Channel Mixer} (4 folds) & 96  &     \textbf{0.227} &     \textbf{0.102} &  \textbf{0.407} &     \textbf{0.399} &           0.348 &  \underline{0.292} &           0.348 &     \textbf{0.291} &  \underline{0.259} &  \textbf{0.171} &    \textbf{0.216} &     \textbf{0.164} \\
& 192 &  \underline{0.328} &  \underline{0.209} &  \textbf{0.427} &     \textbf{0.436} &           0.391 &  \underline{0.358} &           0.371 &     \textbf{0.327} &              0.297 &           0.227 &                    \textbf{0.252} &     \textbf{0.205} \\
& 384 &              0.482 &              0.432 &  \textbf{0.445} &              0.464 &           0.415 &              0.374 &           0.395 &     \textbf{0.367} &              0.345 &           0.297 &                    \textbf{0.298} &     \textbf{0.268} \\
\midrule
Pre-trained Univariate \texttt{MT} & 96  &              0.236 &              0.109 &           0.413 &              0.410 &           0.348 &              0.296 &           0.348 &              0.295 &              0.260 &           0.173 &                      0.222 &              0.173 \\
& 192 &              0.338 &              0.217 &           0.432 &              0.442 &           0.390 &              0.359 &           0.369 &              0.331 &              0.296 &           0.226 &                             0.258 &              0.217 \\
                                  & 384 &     \textbf{0.479} &     \textbf{0.425} &  \textbf{0.445} &     \textbf{0.459} &  \textbf{0.408} &              0.374 &           0.393 &              0.373 &     \textbf{0.339} &  \textbf{0.291} &                             0.301 &              0.277 \\
\bottomrule
\end{tabular}
}
\vspace{0pt}
\caption{MAE and MSE for various models across different datasets and horizons. The best metric for each dataset and horizon is highlighted, results better than \texttt{MOMENT-Tiny} are underlined. \texttt{MT} means \texttt{MOMENT-Tiny} as a backbone architecture, each model is pre-trained on \textbf{ONE Epoch} of the Time Series Pile.}
\label{tab:metrics_forecast_pretrained_full}
\end{table}
\begin{table}[h!]
\centering
\resizebox{\textwidth}{!}{
\begin{tabular}{r|c|cc|cc|cc|cc|cc|cc}
\toprule
\textbf{Model name} & \textbf{Horizon} & \multicolumn{2}{c}{\texttt{Exchange}} & \multicolumn{2}{c}{\texttt{ETTh1}} & \multicolumn{2}{c}{\texttt{ETTh2}} & \multicolumn{2}{c}{\texttt{ETTm1}} & \multicolumn{2}{c}{\texttt{ETTm2}} & \multicolumn{2}{c}{\texttt{Weather}} \\
& & MAE & MSE & MAE & MSE & MAE & MSE & MAE & MSE & MAE & MSE & MAE & MSE \\
\midrule 
MT + Graph Transformer layer & 96  &              0.210 &              0.089 &              0.409 &     \textbf{0.382} &              0.365 &              0.316 &  \underline{0.349} &              0.304 &              0.256 &              0.170 &  \textbf{0.197} &  \underline{0.151} \\
              & 192 &              0.335 &              0.222 &              0.436 &              0.435 &              0.413 &              0.385 &  \underline{0.372} &              0.339 &  \underline{0.299} &  \underline{0.231} &  \textbf{0.241} &  \underline{0.196} \\
              & 384 &  \underline{0.470} &  \underline{0.409} &              0.477 &              0.489 &              0.429 &              0.399 &  \underline{0.397} &  \underline{0.376} &              0.354 &              0.310 &  \textbf{0.292} &              0.264 \\
\midrule
NBeats & 96  &              0.316 &              0.173 &              0.422 &              0.407 &              0.377 &              0.334 &              0.357 &              0.304 &              0.268 &              0.180 &           0.207 &  \underline{0.152} \\
              & 192 &              0.527 &              0.467 &              0.456 &              0.453 &              0.422 &              0.392 &              0.379 &              0.339 &              0.320 &              0.266 &           0.273 &              0.208 \\
              & 384 &              0.739 &              0.933 &              0.500 &              0.523 &              0.500 &              0.503 &              0.413 &              0.394 &              0.394 &              0.389 &           0.323 &              0.273 \\
\midrule
CrossFormer & 96  &              0.385 &              0.260 &              0.448 &              0.427 &              0.540 &              0.575 &              0.365 &              0.317 &              0.383 &              0.312 &           0.218 &     \textbf{0.148} \\
              & 192 &              0.606 &              0.644 &              0.508 &              0.515 &              0.573 &              0.640 &              0.380 &              0.353 &              0.589 &              0.632 &           0.281 &              0.204 \\
              & 384 &              0.704 &              0.774 &              0.668 &              0.770 &              0.644 &              0.746 &              0.507 &              0.501 &              0.573 &              0.601 &           0.378 &              0.329 \\
\midrule
iTransformer & 96  &              0.231 &              0.104 &              0.408 &              0.391 &              0.373 &              0.325 &              0.361 &              0.311 &              0.268 &              0.186 &           0.207 &              0.160 \\
              & 192 &              0.332 &              0.208 &              0.434 &              0.434 &              0.417 &              0.400 &              0.383 &              0.350 &              0.315 &              0.258 &           0.250 &              0.206 \\
              & 384 &  \underline{0.480} &  \underline{0.423} &              0.452 &              0.462 &              0.434 &              0.414 &              0.409 &              0.397 &              0.353 &              0.310 &           0.299 &              0.272 \\
 \midrule
 Infini-Channel Mixer MT (Ours) & 96  &              0.219 &              0.096 &     \textbf{0.400} &  \underline{0.383} &     \textbf{0.346} &     \textbf{0.295} &  \underline{0.348} &     \textbf{0.298} &              0.255 &              0.170 &           0.203 &  \underline{0.154} \\
              & 192 &     \textbf{0.316} &     \textbf{0.193} &     \textbf{0.421} &     \textbf{0.420} &     \textbf{0.396} &  \underline{0.368} &     \textbf{0.369} &     \textbf{0.330} &  \underline{0.296} &  \underline{0.229} &           0.244 &  \underline{0.197} \\
              & 384 &     \textbf{0.467} &     \textbf{0.407} &     \textbf{0.440} &              0.446 &     \textbf{0.417} &     \textbf{0.385} &     \textbf{0.394} &     \textbf{0.372} &              0.345 &              0.304 &           0.295 &  \underline{0.263} \\
\midrule
Infini-Channel Mixer + Static MT (Ours) & 96  &              0.224 &              0.100 &  \underline{0.405} &              0.390 &              0.354 &              0.300 &     \textbf{0.346} &     \textbf{0.298} &     \textbf{0.252} &     \textbf{0.167} &           0.201 &     \textbf{0.148} \\
              & 192 &  \underline{0.322} &              0.204 &              0.428 &              0.427 &  \underline{0.398} &     \textbf{0.362} &     \textbf{0.369} &     \textbf{0.330} &     \textbf{0.294} &     \textbf{0.224} &           0.244 &     \textbf{0.192} \\
              & 384 &  \underline{0.483} &  \underline{0.437} &  \underline{0.441} &              0.448 &              0.438 &              0.409 &  \underline{0.395} &  \underline{0.373} &              0.346 &              0.303 &           0.293 &     \textbf{0.258} \\
\midrule
Channel Concatenation & 96  &              0.223 &              0.099 &              0.410 &              0.393 &              0.356 &              0.303 &              0.353 &              0.302 &              0.264 &              0.180 &           0.211 &              0.166 \\
              & 192 &              0.329 &              0.213 &              0.430 &              0.429 &     \textbf{0.396} &  \underline{0.370} &              0.377 &              0.339 &              0.305 &              0.238 &           0.251 &              0.210 \\
              & 384 &  \underline{0.478} &  \underline{0.418} &              0.448 &              0.456 &              0.428 &              0.399 &              0.400 &              0.380 &              0.364 &              0.329 &           0.297 &              0.273 \\
\midrule
Univariate MT & 96  &     \textbf{0.204} &     \textbf{0.084} &              0.407 &              0.388 &              0.354 &              0.300 &              0.351 &              0.300 &              0.254 &     \textbf{0.167} &           0.201 &              0.155 \\
              & 192 &              0.323 &              0.202 &              0.425 &              0.422 &              0.399 &              0.376 &              0.374 &              0.338 &              0.303 &              0.235 &           0.242 &              0.198 \\
              & 384 &              0.499 &              0.460 &              0.442 &     \textbf{0.445} &              0.420 &              0.399 &              0.399 &              0.379 &     \textbf{0.344} &     \textbf{0.301} &  \textbf{0.292} &              0.264 \\

\bottomrule
\end{tabular}
}
\vspace{0pt}
\caption{MAE and MSE for various models, forecasting, supervised setup, across different datasets and horizons. The best metric for each dataset and horizon is highlighted, results better than \texttt{MOMENT-Tiny} are underlined. MT means \texttt{MOMENT-Tiny} as a backbone architecture. Our method shows promising results, with notably the moment backbone paired with the Graph Transformer layer showing good results on the Weather dataset. }
\label{tab:metrics_forecast_supervised_full}
\end{table}

\subsection{Classification}
In the multivariate case, \texttt{MOMENT-Tiny + Infini-Channel Mixer} achieves comparable results to \texttt{MOMENT-Tiny} \ref{table:summary_stats}, although mean and median accuracy is higher, the number of wins/losses for \texttt{MOMENT-Tiny + Infini-Channel Mixer} is 9155 / 6512, which is worse than for \texttt{MOMENT-Tiny} 9497 / 6170. We hypothesize that this happens because the $\beta$ parameters allow the model backbone to adjust during the fine-tuning stage. In the univariate case, our model with Channel-Mixing performs worse than the vanilla one. This points towards the hypothesis that meaningful filtering of the aggregated memory is an important direction of future work. \ref{table:multivariate_classification}
\ref{table:univariate_classification}.
\begin{longtable}{lcc}
\toprule
Dataset & \texttt{MOMENT-Tiny} & \texttt{MOMENT-Tiny + Infini-Channel Mixing} Mixer \\
\midrule
\endfirsthead

\toprule
Dataset & \texttt{MOMENT-Tiny} &\texttt{MOMENT-Tiny + Infini-Channel Mixing} Mixer \\
\midrule
\endhead

\midrule
\multicolumn{3}{r}{\textit{Continued on next page}} \\
\midrule
\endfoot
\bottomrule
\endlastfoot
ArticularyWordRecognition & 0.923 & \textbf{0.943} \\
AtrialFibrillation & \textbf{0.400} & 0.333 \\
BasicMotions & 0.550 & \textbf{0.875} \\
Cricket & \textbf{0.972} & \textbf{0.972} \\
DuckDuckGeese & \textbf{0.540} & 0.340 \\
EigenWorms & \textbf{0.557} & 0.542 \\
Epilepsy & 0.949 & \textbf{0.971} \\
ERing & \textbf{0.852} & 0.811 \\
EthanolConcentration & \textbf{0.346} & 0.281 \\
FingerMovements & 0.500 & \textbf{0.650} \\
HandMovementDirection & 0.230 & \textbf{0.284} \\
Handwriting & \textbf{0.212} & 0.179 \\
Heartbeat & \textbf{0.722} & \textbf{0.722} \\
JapaneseVowels & 0.676 & \textbf{0.692} \\
Libras & \textbf{0.850} & 0.822 \\
LSST & \textbf{0.329} & 0.266 \\
MotorImagery & 0.480 & \textbf{0.520} \\
NATOPS & 0.772 & \textbf{0.828} \\
PEMS-SF & \textbf{0.879} & 0.855 \\
PenDigits & 0.959 & \textbf{0.971} \\
PhonemeSpectra & \textbf{0.212} & 0.175 \\
RacketSports & 0.697 & \textbf{0.717} \\
SelfRegulationSCP1 & \textbf{0.785} & 0.724 \\
SelfRegulationSCP2 & 0.489 & \textbf{0.550} \\
SpokenArabicDigits & \textbf{0.966} & 0.928 \\
StandWalkJump & 0.400 & \textbf{0.467} \\
\bottomrule
\\
\caption{Accuracy for multivariate datasets (UCR/UEA) classification for \texttt{MOMENT-Tiny} and \texttt{MOMENT-Tiny + Infini-Channel Mixing}.}
\label{table:multivariate_classification}
\end{longtable}
\begin{longtable}{lccc}
\toprule
Dataset & \texttt{MOMENT-Tiny} & \texttt{MOMENT-Tiny + Infini-Channel Mixing} \\
\midrule
\endfirsthead

\toprule
Dataset & \texttt{MOMENT-Tiny} & \texttt{MOMENT-Tiny + Infini-Channel Mixing} \\
\midrule
\endhead

\midrule
\multicolumn{3}{r}{\textit{Continued on next page}} \\
\midrule
\endfoot
\bottomrule
\endlastfoot
GestureMidAirD2 & \textbf{0.538} & 0.515 \\
UWaveGestureLibraryX & \textbf{0.790} & 0.745 \\
GesturePebbleZ2 & \textbf{0.791} & 0.722 \\
ECG5000 & 0.921 & \textbf{0.934} \\
OSULeaf & \textbf{0.802} & 0.719 \\
MedicalImages & \textbf{0.759} & 0.658 \\
Ham & 0.543 & \textbf{0.581} \\
DistalPhalanxTW & 0.612 & \textbf{0.619} \\
ProximalPhalanxOutlineCorrect & \textbf{0.842} & 0.838 \\
FreezerRegularTrain & \textbf{0.961} & 0.946 \\
TwoLeadECG & 0.748 & \textbf{0.838} \\
GunPointMaleVersusFemale & 0.975 & \textbf{0.984} \\
Trace & \textbf{0.980} & 0.940 \\
SmoothSubspace & 0.793 & \textbf{0.867} \\
MiddlePhalanxTW & \textbf{0.558} & 0.552 \\
SyntheticControl & \textbf{0.950} & 0.880 \\
ShapesAll & \textbf{0.772} & 0.725 \\
AllGestureWiimoteX & \textbf{0.627} & 0.520 \\
Wafer & \textbf{0.996} & 0.994 \\
FaceFour & \textbf{0.557} & 0.443 \\
CricketX & \textbf{0.641} & 0.541 \\
DistalPhalanxOutlineCorrect & \textbf{0.714} & 0.707 \\
ChlorineConcentration & 0.671 & \textbf{0.702} \\
Chinatown & \textbf{0.980} & 0.971 \\
GestureMidAirD1 & \textbf{0.623} & 0.523 \\
MiddlePhalanxOutlineAgeGroup & \textbf{0.526} & 0.468 \\
UMD & \textbf{0.965} & 0.917 \\
Crop & \textbf{0.701} & 0.701 \\
GesturePebbleZ1 & \textbf{0.901} & 0.797 \\
WordSynonyms & \textbf{0.577} & 0.475 \\
ArrowHead & \textbf{0.571} & 0.520 \\
Wine & \textbf{0.556} & 0.500 \\
Coffee & \textbf{0.821} & 0.536 \\
Earthquakes & \textbf{0.748} & \textbf{0.748} \\
Herring & \textbf{0.594} & \textbf{0.594} \\
Beef & 0.667 & \textbf{0.700} \\
MiddlePhalanxOutlineCorrect & 0.526 & \textbf{0.591} \\
ECGFiveDays & \textbf{0.844} & 0.742 \\
Yoga & 0.742 & \textbf{0.751} \\
Adiac & \textbf{0.627} & 0.619 \\
MoteStrain & 0.654 & \textbf{0.712} \\
Strawberry & 0.922 & \textbf{0.924} \\
InsectWingbeatSound & \textbf{0.563} & 0.508 \\
DodgerLoopWeekend & \textbf{0.848} & 0.703 \\
Meat & \textbf{0.833} & \textbf{0.833} \\
MelbournePedestrian & 0.870 & \textbf{0.872} \\
FaceAll & \textbf{0.666} & 0.615 \\
FacesUCR & \textbf{0.652} & 0.606 \\
AllGestureWiimoteY & \textbf{0.671} & 0.589 \\
ShakeGestureWiimoteZ & \textbf{0.760} & 0.720 \\
BME & 0.947 & \textbf{0.960} \\
FordB & \textbf{0.810} & 0.786 \\
Fish & \textbf{0.777} & 0.629 \\
SonyAIBORobotSurface2 & \textbf{0.821} & 0.753 \\
FiftyWords & \textbf{0.677} & 0.589 \\
ToeSegmentation1 & \textbf{0.925} & 0.851 \\
FreezerSmallTrain & 0.744 & \textbf{0.747} \\
TwoPatterns & \textbf{0.964} & 0.821 \\
ShapeletSim & \textbf{0.650} & 0.594 \\
Plane & \textbf{0.952} & 0.933 \\
GestureMidAirD3 & \textbf{0.315} & 0.292 \\
DiatomSizeReduction & \textbf{0.784} & 0.765 \\
CricketZ & \textbf{0.669} & 0.564 \\
Lightning7 & \textbf{0.589} & 0.548 \\
UWaveGestureLibraryY & \textbf{0.723} & 0.633 \\
GunPointAgeSpan & 0.949 & \textbf{0.965} \\
DistalPhalanxOutlineAgeGroup & \textbf{0.683} & 0.676 \\
SwedishLeaf & \textbf{0.875} & 0.840 \\
CBF & \textbf{0.864} & 0.671 \\
BeetleFly & \textbf{0.750} & 0.650 \\
AllGestureWiimoteZ & \textbf{0.569} & 0.466 \\
DodgerLoopDay & \textbf{0.400} & 0.263 \\
GunPointOldVersusYoung & \textbf{0.933} & 0.889 \\
FordA & \textbf{0.934} & 0.913 \\
ItalyPowerDemand & \textbf{0.935} & 0.900 \\
ProximalPhalanxOutlineAgeGroup & \textbf{0.839} & 0.829 \\
GunPoint & \textbf{0.960} & 0.940 \\
ProximalPhalanxTW & \textbf{0.737} & 0.732 \\
PickupGestureWiimoteZ & \textbf{0.640} & 0.480 \\
SonyAIBORobotSurface1 & 0.626 & \textbf{0.687} \\
PowerCons & \textbf{0.894} & 0.833 \\
PhalangesOutlinesCorrect & \textbf{0.699} & 0.650 \\
BirdChicken & \textbf{0.900} & 0.750 \\
ToeSegmentation2 & \textbf{0.938} & 0.869 \\
CricketY & \textbf{0.585} & 0.482 \\
ElectricDevices & 0.634 & \textbf{0.638} \\
DodgerLoopGame & \textbf{0.623} & 0.551 \\
Fungi & \textbf{0.887} & 0.806 \\
Symbols & \textbf{0.857} & 0.842 \\
UWaveGestureLibraryZ & \textbf{0.750} & 0.678 \\
ECG200 & \textbf{0.840} & \textbf{0.840} \\
\bottomrule
\\
\caption{Accuracy for univariate datasets (UCR/UEA) classification for \texttt{MOMENT-Tiny} and \texttt{MOMENTiTiny + Infini-Channel Mixing}}
\label{table:univariate_classification}
\end{longtable}

\end{document}